\pdfoutput=1

\documentclass[11pt]{article}
\usepackage{fancyhdr}

\usepackage[]{EMNLP2023}
\usepackage{multirow}
\usepackage{booktabs}

\usepackage{times}
\usepackage{latexsym}
\usepackage{multirow}
\usepackage{algorithm}
\usepackage{algorithmic}
\usepackage{hyperref}
\usepackage{adjustbox}
\usepackage{amsmath}
\usepackage{amssymb}
\usepackage{caption}
\usepackage{fancyhdr}

\usepackage[T1]{fontenc}

\usepackage[utf8]{inputenc}

\usepackage[activate=true,
final,
tracking=false,   
kerning=true,
spacing=true,
factor=1050,
stretch=30,
shrink=30]{microtype}      

\usepackage{todonotes}
\makeatletter
\newcommand*\iftodonotes{\if@todonotes@disabled\expandafter\@secondoftwo\else\expandafter\@firstoftwo\fi}  
\makeatother

\fancypagestyle{mystyle}{%
    \fancyhf{} 
    \fancyfoot[C]{\thepage} 
}

\newcommand{\E}{\mathbb{E}}
\def\rvx{{\mathbf{x}}}
\newcommand{\pdata}{p_{\rm{data}}}

\usepackage{inconsolata}



\title{UT5: Pretraining Non autoregressive T5 with unrolled denoising}

\author{Mahmoud G. Salem, Jiayu Ye, Chu-Cheng Lin, Frederick Liu \\
        Google  \\ \{* \}@google.com}

\begin{document}


\maketitle
\begin{abstract}

Recent advances in Transformer-based Large Language Models have made great strides in natural language generation. However, to decode $K$ tokens, an autoregressive model needs $K$ sequential forward passes, which may be a performance bottleneck for large language models. 
Many non-autoregressive (NAR) research are aiming to address this sequentiality bottleneck, albeit many have focused on a dedicated architecture in supervised benchmarks. 
In this work, we studied unsupervised pretraining for non auto-regressive T5 models via unrolled denoising and shown its SoTA results in downstream generation tasks such as SQuAD question generation and XSum.

\end{abstract}

\begin{figure*} \
  \centering
  \includegraphics[width=0.7\textwidth]{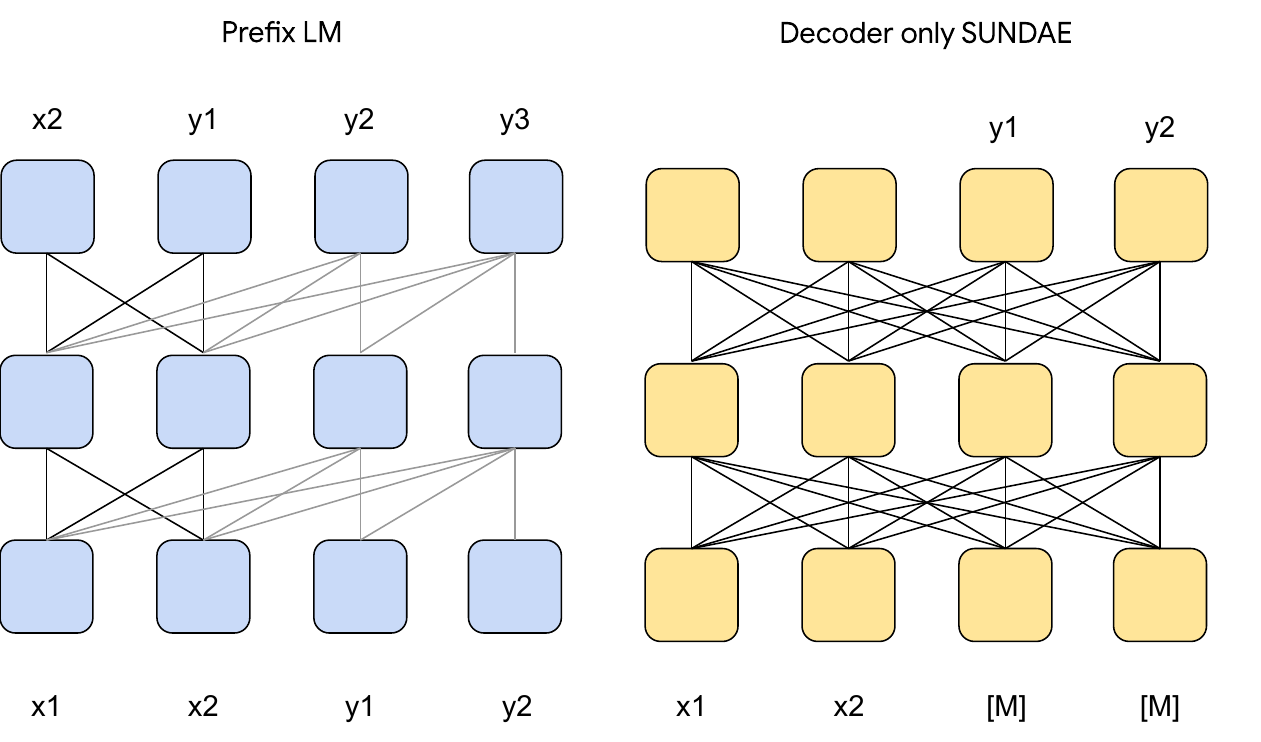}
  \caption{ Illustration of prefix Language Model versus Decoder-only bidirectional de-noising model. }
  \label{decoder-only-illustration}
\end{figure*}

\section{Introduction}

Large language models (LLMs) have revolutionized natural language processing (NLP) by enabling automatic text generation and prediction. Traditionally, language models are \emph{autoregressive}: they generate a sequence of tokens one by one, conditioning each token on the previously generated ones.
While this approach has led to impressive results \cite{openai2023gpt4, anil2023palm}, it suffers from slow inference due to its sequential nature. Several studies \cite{gu2018nonautoregressive, ghazvininejad2019maskpredict}  have explored the use of non-autoregressive generation for language modeling, where tokens can be generated in parallel, without the need of conditioning on previously generated ones. Non-autoregressive generation has shown promising results in terms of efficiency and speed, and has the potential to be applied to various NLP tasks \cite{liu2020glge}. Pretraining has proven the foundational procedure for autoregressive generation \cite{devlin-etal-2019-bert, radford2018improving}. However, few studies have focused on pretraining for non-autoregressive language modeling for efficient language generation .  The main advantage of non-autoregressive generation is parallel generation of all tokens, making it faster than auto-regressive generation. However, non-autoregressive generation usually exhibits quality gaps when comparing with similar sized autoregressive models \cite{gu2020fully}.

In this paper, we propose a pretraining regime to improve the quality of non-autoregressive generation. To explore the effects of pretraining on decoder-only models, we employed step-unrolled denoising \cite{savinov2021step} to pretrain the models.  In the rest of the paper, we describe our proposed pretraining regime in detail and evaluate its effectiveness in improving the quality of efficientnon-autoregressive text generation.

Our contributions are:
\begin{itemize}
\item Introduce training regime for non-autoregressive models for efficient language generation,
\item We show that the non-autoregressive pretraining with unrolled denoising significantly improves the results on downstream benchmarks compared to training from scratch. 
\item We are able to achieve SoTA results on downstream evaluations with similar parameter count. 
\end{itemize}

\section{Related work}

Pretraining language models on large-scale data has shown great success for auto-regressive language models \cite{devlin2018bert,ilic2018deep,radford2018improving}. The models are pre-trained on large-scale data in a self-supervised manner then finetuned on downstream tasks like text classification and machine translation.  While pre-training is a standard in many autoregressive language tasks, it is understudied in non-autoregressive settings.  Some efforts have been made to study and adapt pre-training for non auto-regressive models. \citep{guo2020incorporating} incorporates two BERT models into machine translation using mask-predict decoding method, their method utilizes two pre-trained BERT models one as the encoder and one as the decoder, and then inserts adapter layers into each layer. \citep{su2021non} follows similar regime but uses one  BERT as the backbone model and then add a CRF output layer which captures the target side dependency and improves the performance. Further \citep{li2022universal}  introduced CeMAT which uses a bidirectional encoder and decoder architecture. The model is jointly trained with Masked Language modeling (MLM) for the decoder and Conditional Masked Language Modeling (CMLM) for the decoder with a cross attention module for bridging them. The model seeks to enhance multilingual ability in machine translation by pre-training on large-scale monolingual and bilingual texts in many languages and using an aligned code-switching strategy than finetuned on NAT and AT tasks.

\textbf{SUNDAE} \cite{savinov2021step} is a novel method for training denoising models for text generation. SUNDAE improves upon traditional denoising autoencoders by unrolling the decoding process for multiple steps and adding noise at each step. resulting in a more robust and effective model for generating text. The authors demonstrated the effectiveness of the SUNDAE method in several text generation tasks, including sentence completion and language modeling, and showed that it outperformed other state-of-the-art methods in terms of both quality and efficiency. The SUNDAE method provides a promising approach to text generation and has practical applications in various natural language processing tasks. However, SUNDAE language generation suffers a huge drop in performance when adapted in non-auto-regressive generation setting. In this study we focus on recovering the drop in performance using large-scale pretraining. 

\textbf{BANG} \cite{qi2021bang} investigated pretraining an LLM using a mixture of autoregressive and non-autoregressive objective functions. Their downstream tasks include machine translation, summarization, and dialogue generation. BANG achieves state-of-the-art performance on several benchmark datasets, demonstrating the effectiveness of large-scale pretraining for bridging the gap between autoregressive and non-autoregressive language generation. We consider the BANG model to be a potential baseline, where the non-autoregressive parametrization simply dropped conditioning on previously generated tokens.


\section{Method}\label{sec:t2t}
Pretraining techniques such as masked language modeling (MLM) on large-scale data have shown to be effective in improving the performance of neural language models. In this section, we investigate the effects of large-scale pretraining on decoder-only non-autoregressive models. We adopted SUNDAE \citep{savinov2021step}, a two-step training method for generative modeling of discrete sequences using denoising autoencoders and Markov chain models. The training process includes unrolled denoising, which involves starting the chain from corrupted data samples instead of the prior distribution. The model learns to denoise samples that it is likely to encounter during full unrolling used at sample time.

\begin{equation} \label{eq:logits2}
    L^{(t)}(\theta) :=
    -\E\!\!\!\!_{\substack{
        \rvx\sim\pdata\\
        \rvx_0\sim q(\cdot|\rvx) \\
        \rvx_1\sim f_{\theta}(\cdot|\rvx_0) \\
        }}\!\!\!\![\log f_{\theta}(\rvx|\rvx_{i})],
\end{equation}
where $\rvx_i$ is the $i$th iteration denoised result, $q(\cdot|\rvx)$ is the corruption function, and $f_{\theta}$ is the network.

We investigate the effect of pretraining on the decoder-only architecture proposed in \cite{radford2018improving} combined with SUNDAE two-step training procedure as our baseline model. The pretraining is done on the Colossal Clean Crawled Corpus (C4) dataset. The pretraining objective is similar to prefix language modeling but with bidirectional attention as shown in Figure \ref{decoder-only-illustration}. Following pretraining, we finetune the model on several downstream tasks. 

\subsection{Model Details}

We ground the work on T5 base \cite{raffel2020exploring} and develop a decoder-only model on top. Our baseline model utilizes a decoder-only transformer-based architecture with bidirectional self-attention. Specifically, we employ a 12-layer decoder with hidden states of dimension 768. This is comparable with BANG with 6 layers of encoder and 6 layers of decoder with the same hidden dimension. 

Several NAR techniques \cite{gu2018nonautoregressive, savinov2021step} try to incorporate the output sentence length information during the training allowing NAR models to have some approximate of the output length. To keep our study simple and focused on the value of pretraining, we omit the use of length prediction neither as an auxiliary loss or a separate module. Alternatively, the model is trained to predict padding tokens to fill the target sequence buffer.

\subsection{Training Strategy}

During the pretraining phase, our model underwent training for 1 million steps on the C4 dataset with a batch size of 128 and
a sequence length of 512 inputs and 114 targets. We explore span corruption and prefix LM strategies during pretraining while observing the latter is more stable. One of the hypothesis is a single span corruption target is shorter hence less meaningful to unroll. Hence for the studies below, we use Prefix LM objective with bidirectional attention (Figure.\ref{decoder-only-illustration}). This process allowed the model to develop a comprehensive understanding of language patterns and contextual relationships.

For the subsequent finetuning stage, the model is fine-tuned on a specific downstream task for 50k steps, employing a learning rate of 0.0001. The pretraing helps the model to efficiently finetune on different downstream tasks with fewer number of steps.  The finetuning process further refined the model's parameters and enabled it to adapt to the nuances and requirements of the target task. During the model inference evaluation, the model unrolls 10 steps from the mask then decodes text as output.   

\begin{table*}[ht]
\centering

\scalebox{0.75}{
\begin{tabular}{c|c|cccc|ccc}
\toprule[1.2pt]
& & \multicolumn{4}{c|}{XSum} & \multicolumn{2}{c}{SQuAD}  & \\
 Model & Pretrain &  ROUGE-1 & ROUGE-2 & ROUGE-L & OVERALL & ROUGE-L & BLEU-4 \\ 
 \midrule \midrule
NAT \cite{gu2018nonautoregressive} & No & 24.04 & 3.88 & 20.32& 16.08 & 31.51 & 2.46 \\
iNAT \cite{lee2018deterministic} & No & 24.02 &  3.99 & 20.36 & 16.12 & 32.44 & 2.33 \\
CMLM \cite{ghazvininejad2019mask} & No &   23.82 &  3.60 & 20.15 & 15.86 &  31.58  &  2.51 \\
LevT \cite{gu2019levenshtein} & No  &  24.75 & 4.18 & 20.87 & 16.60 & 31.38 & 2.27 \\
BANG NAR \cite{qi2021bang} & Yes  & 32.59 & 8.98 & 27.41 & 22.99 & 44.07 & 12.75 \\
BANG semi-NAR & Yes & 34.71 & 11.71 & 29.16 & 25.19 & \textbf{47.39} & \textbf{17.62} \\
\midrule
Ours (no prefix-lm pretraining)& No    & 32.56 & 11.8	 & 26.17 & 23.51  & 31.36 & 	3.903 \\
Ours (with prefix-lm pretraining) & Yes & \textbf{35.80} & \textbf{14.03} & \textbf{29.27} &  \textbf{26.36} & 45.75 & 12.47 \\

\bottomrule[1.2pt]
\end{tabular}}
\caption{NAR results on the XSum and SQuAD 1.1 question generation.\\}
\label{tab:xsum_squad}
\end{table*}

\section{Experiments}
We conduct the experiments to study the effect of pretraining on decoder-only NAR models. We analyze the performance on these models on downstream tasks with and without pretraining. Our experiments are all conducted through JAX/Flax \cite{bradbury2018jax} using the T5x framework \cite{roberts2022scaling}. We use TPU-v3 chips for pretraining and finetuning, typical pretraining jobs use 256 chips for a week and finetuning jobs use 16 to 64 chips for a day.

\subsection{Datasets}
\label{sec:glue}
\textbf{Pretraining.} For our pretraining experiments, we use the C4 dataset, which is a large-scale web document corpus created by scraping the Common Crawl data. The C4 dataset contains over 750GB of text data and includes a diverse range of topics, such as news, blogs, and online forums. The text data in the C4 dataset is preprocessed and tokenized into individual sentences, making it suitable for language modeling tasks. The C4 dataset has several advantages over other datasets for pretraining, such as its large size and diversity. The size of the dataset allows for the training of large-scale language models, which have been shown to achieve state-of-the-art performance on various NLP tasks. Additionally, the diversity of the C4 dataset helps to capture the different styles and registers of language used in the web documents, making the pretraining models more robust to different text domains.

To evaluate our approach, we conduct experiments on following two popular generation benchmarks for downstream evaluation:

\textbf{XSum.} The XSum dataset \cite{narayan2018don} contains over 227,000 news articles and their corresponding summaries from the British Broad- casting Corporation (BBC). The articles are taken from a wide range of topics, such as politics, business, sports, and entertainment. The summaries are written to capture the main idea and salient points of the articles in a single sentence. The average input and output lengths are 358.5 and 21.1, respectively.

\textbf{SQuAD 1.1} \cite{rajpurkar2016squad}  is a popular benchmark dataset for evaluating the performance of question answering models. It was released by Stanford University in 2016 and contains over 100,000 questions with their corresponding answers, all based on a set of Wikipedia articles. After preprocessing, the dataset contains 98K <answer, passage, question> data triples. Input is formatted as <answer [SEP] passage> following GLGE. The average input and output lengths are 149.4 and 11.5, respectively.

\subsection{Results}
In this section, we show large scale pretraining using prefix-lm leads to huge improvement in performance for NAR decoder-only models. We evaluate our approach on two popular datasets. For XSum dataset, we use a combination of ROUGE score \cite{lin-2004-rouge} to evaluate different models. As shown in table \ref{tab:xsum_squad}, we observe +2.9 ROUGE-L score when the model is pretrained. Also the model outperformed both BANG NAR and Semi-NAR and CMLM in terms of all three ROUGE metrics. We also evaluated our approach on Squad 1.1 question generation task, our model was able to show +14.4 ROUGE-L and +8.6 BLEU-4 when the model is pretrained. And it demonstrates +1.7 ROUGE-L improvement in performance compared to BANG NAR while -2.7 ROULGE-L compared to BANG semi-NAR.

\section{Ablation Studies}

\subsection{Model Architecture}

\begin{table}[ht]
\centering
\scalebox{0.8}{
\begin{tabular}{c|ccc}
 Model & @500k & @1M & best  \\ 
\hline
Decoder only Pretrained & 21.6 & 21.76 & 21.76 \\
Encoder Decoder Pretrained  & 20.13 & 18.42 &	21.73 \\

\end{tabular}}
\caption{BLEU on WMT14 EN$\to$DE.}
\label{tab:arch-wmt}
\end{table}

We conduct preliminary experiments on WMT14 using EN-DE on both encoder-decoder and decoder only model. The max BLEU number for encoder-decoder and decoder only model have negligible difference while the encoder-decoder model has a high variance during eval. Hence we utilize the decoder only architecture for the main study on other downstream benchmarks. 

\subsection{Sample Efficiency}

\begin{table}[ht]
\centering
\scalebox{0.8}{
\begin{tabular}{c|cc}
 Model & @500k & @1M \\ 
\hline
Decoder only From scratch & 14.57 & 21.89 \\
Decoder only Pretrained & 21.6 & 21.76\\

\end{tabular}}
\caption{Decoder-only BLEU the WMT14 EN$\to$DE.}
\label{tab:de-wmt}
\end{table}

In Table \ref{tab:de-wmt}, we present the WMT14 ENDE numbers for pretrained vs from scratch numbers. We see although the final numbers have negligible difference, the pretrained model is more sample efficient, reaching higher number with the same fine-tune steps. Note that this number is not comparable with SoTA WMT results because of the length predictor, for fair comparison, please refer to SUNDAE Appendix Figure 4a. 

\section{Conclusion and Future Work}\label{sec:conclusion}

In this work, we investigate the effect of pretraining for non-autoregressive decoder only SUNDAE. We show that pretraining should be considered a foundational block for non-autoregressive model. For future work, there is a natural question: Will the non-autoregressive model scales with data size and model parameters as larger autoregressive models do.


%




\bibliography{anthology,emnlp2022}
\bibliographystyle{acl_natbib}

\appendix

\end{document}